\def\year{2022}\relax
\documentclass[letterpaper]{article} 
\usepackage{aaai22}  
\usepackage{times}  
\usepackage{helvet}  
\usepackage{courier}  
\usepackage[hyphens]{url}  
\usepackage{graphicx} 
\usepackage{natbib}  
\usepackage{caption} 
\usepackage{booktabs} 
\usepackage{warpcol}
\DeclareCaptionStyle{ruled}{labelfont=normalfont,labelsep=colon,strut=off} 
\usepackage{algorithm}
\usepackage{algorithmic}
\usepackage{amsmath}
\usepackage{amssymb}
\usepackage{multirow}
\usepackage{color}
\usepackage{pifont}
\usepackage{newfloat}
\usepackage{listings}
\floatstyle{ruled}
\newfloat{listing}{tb}{lst}{}
\floatname{listing}{Listing}
\title{MODNet: Real-Time Trimap-Free Portrait Matting via Objective Decomposition}
\author{
    Zhanghan Ke\textsuperscript{\rm 1,2},\;
    Jiayu Sun\textsuperscript{\rm 1},\;
    Kaican Li\textsuperscript{\rm 2},\;
    Qiong Yan\textsuperscript{\rm 2},\;
    Rynson W.H. Lau\textsuperscript{\rm 1}
}
\affiliations{
    \textsuperscript{\rm 1}Department of Computer Science, City University of Hong Kong \;\;\;\;\;\;
    \textsuperscript{\rm 2}SenseTime Research\\
     \{zhanghake2-c,jiayusun5-c\}@my.cityu.edu.hk, \{likaican,yanqiong\}@sensetime.com, rynson.lau@cityu.edu.hk
     
}

\usepackage{bibentry}

\begin{document}

\maketitle

\begin{abstract}
Existing portrait matting methods either require auxiliary inputs that are costly to obtain or involve multiple stages that are computationally expensive, making them less suitable for real-time applications. 
In this work, we present a light-weight matting objective decomposition network (MODNet) for portrait matting in real-time with a single input image. 
The key idea behind our efficient design is by optimizing a series of sub-objectives simultaneously via explicit constraints. 
In addition, MODNet includes two novel techniques for improving model efficiency and robustness.
First, an Efficient Atrous Spatial Pyramid Pooling (e-ASPP) module is introduced to fuse multi-scale features for semantic estimation. Second, a self-supervised sub-objectives consistency (SOC) strategy is proposed to adapt MODNet to real-world data to address the domain shift problem common to trimap-free methods.
MODNet is easy to be trained in an end-to-end manner. It is much faster than contemporaneous methods and runs at 67 frames per second on a 1080Ti GPU. Experiments show that MODNet outperforms prior trimap-free methods by a large margin on both Adobe Matting Dataset and a carefully designed photographic portrait matting (PPM-100) benchmark proposed by us. Further, MODNet achieves remarkable results on daily photos and videos. Our code and models are available at: {\color{blue} https://github.com/ZHKKKe/MODNet}, and the PPM-100 benchmark is released at: {\color{blue} https://github.com/ZHKKKe/PPM}.
\end{abstract}
\begin{figure*}[t]
    \begin{center}
       \includegraphics[width=0.99\linewidth]{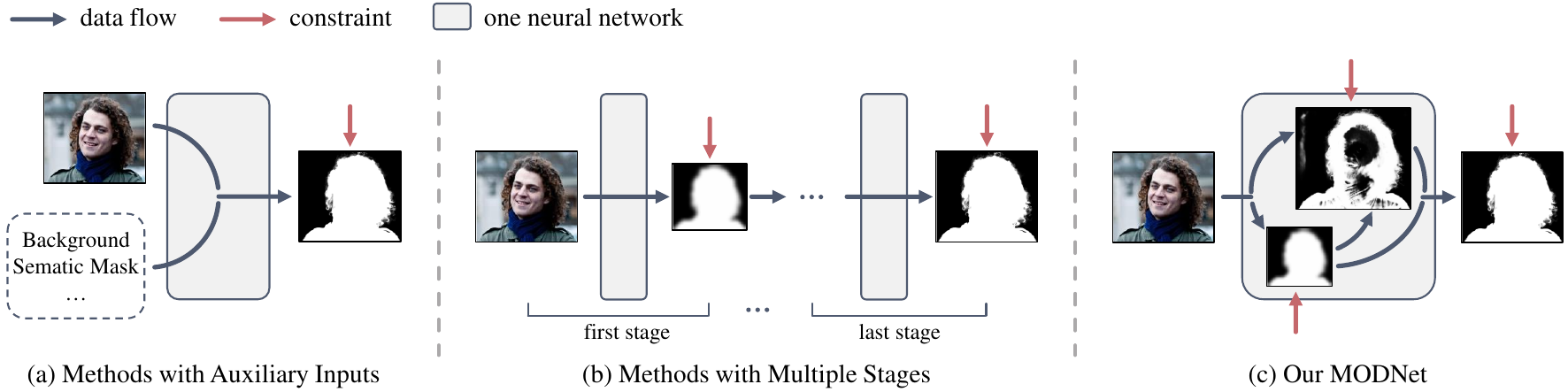}
    \end{center}
    \caption{\textbf{Different Trimap-free Matting Approaches.} 
    Existing trimap-free matting methods either (a) require auxiliary inputs to address the complex matting objective directly or (b) consist of multiple stages to address the matting sub-objectives sequentially. Both of them are less suitable for real-time applications.
    Instead, (c) our MODNet solves the matting sub-objectives simultaneously with only a single input image, which is more efficient and effective. 
     }
    \label{fig:framework}
\end{figure*}

\section{Introduction}
Portrait matting aims to predict a precise alpha matte that can be used to extract the persons in a given image or video. It has a wide variety of applications, such as photo editing and movie re-creation. 
Most existing matting methods use a pre-defined trimap as a priori to produce an alpha matte \cite{AdaMatting, CAMatting, GCA, IndexMatter, SampleMatting, DIM}.
However, trimaps are costly to annotate. Although a depth sensor \cite{ToF} can ease the task, the resulting trimaps may suffer from low precision. Some recent trimap-free methods attempt to eliminate the model dependence on the trimap. For example, background matting \cite{BM} replaces the trimap by a separate background image. Other methods \cite{SHM, BSHM, DAPM} include multiple stages ({\it i.e.}, with several independent models that are optimized sequentially) to first generate a pseudo trimap or semantic mask, which is then used to serve as the priori for alpha matte prediction.
Nonetheless, using the background image as input has to take and align two photos, while having multiple stages would significantly increase the inference time. These drawbacks make all aforementioned matting methods not suitable for real-time applications, such as camera preview. Besides, limited by insufficient amount of labeled training data, existing trimap-free methods often suffer from the domain shift problem \cite{DomainShift} in practice, {\it i.e.}, the models cannot generalize well to real-world data.

In this work, we present MODNet, a light-weight model for real-time trimap-free portrait matting. As illustrated in Fig.\;\ref{fig:framework}, unlike prior methods which require auxiliary inputs or consist of multiple stages, MODNet inputs a single RGB image and applies explicit constraints to solve matting sub-objectives simultaneously in one stage. There are two insights behind our design.
First, applying explicit constraints to different parts of the model can address portrait matting effectively under a single input image. In contrast, to obtain comparable results, auxiliary inputs would be necessary for the model trained by only one matting constraint.
Second, optimizing sub-objectives simultaneously can further exploit the model capability by sharing intermediate representations. In contrast, training multiple stages sequentially will accumulate the errors from the early stages and magnify them in subsequent stages.
We further propose two novel techniques to improve MODNet's efficiency and robustness, including (1) an Efficient Atrous Spatial Pyramid Pooling (e-ASPP) module for fast multi-scale feature fusion in portrait semantic estimation, and (2) a self-supervised strategy based on sub-objective consistency (SOC) to alleviate the domain shift problem in real-world data.

MODNet has several advantages over previous trimap-free methods. 
First, MODNet is much faster, running at $67$ frames per second ($fps$) on a GTX 1080Ti GPU with an input size of $512\times512$ (including data loading). 
Second, MODNet achieves state-of-the-art results on both open source Adobe Matting benchmark and our newly proposed PPM-100 benchmark.
Third, MODNet can be easily optimized end-to-end as it is a single model instead of a complex pipeline. Finally, MODNet has better generalization ability, due to our SOC strategy. 

Since open-source portrait matting datasets \cite{DAPM, DIM} are typically small and have limited precision, prior works train and validate their models on private datasets of diverse quality and difficulty levels. As a result, it is difficult to conduct a fair evaluation. In this work, we evaluate existing trimap-free methods under the same environment, {\it i.e.}, all models are trained on the same dataset and validated on the portrait images from Adobe Matting Dataset \cite{DIM} and our newly proposed benchmark. Our benchmark is labelled in high quality, and is more diverse than those used in previous works.

In summary, we present a light-weight network architecture, named MODNet, for real-time trimap-free portrait matting. MODNet includes two novel techniques, an e-ASPP module for efficient semantic feature fusion and a self-supervised SOC strategy to generalize MODNet to new data domain. In addition, we have also designed a new validation benchmark for portrait matting. Our code, pre-trained model, and benchmark are released at {\it https://github.com/ZHKKKe/MODNet} and {\it https://github.com/ZHKKKe/PPM}.

\section{Related Works}
\subsection{Image Matting}
The purpose of the image matting task is to extract the desired foreground $F$ from a given image $I$. 
This task predicts an alpha matte with a precise foreground probability value $\alpha$ for each pixel $i$ as:
\begin{equation}
I^{i} = \alpha^{i} \, F^{i} + (1 - \alpha^{i}) \, B^{i} \, ,    
\end{equation}
where $B$ is the background of $I$. This problem is ill-posed since all variables on the right hand side are unknown. Most existing matting methods take a pre-defined trimap as an auxiliary input, which is a mask containing three regions: absolute foreground ($\alpha = 1$), absolute background ($\alpha = 0$), and unknown area ($\alpha = 0.5$). They aim to estimate the foreground probability inside the unknown area only, based on the priori from the other two regions.

Traditional matting algorithms heavily rely on low-level features, {\it e.g.}, color cues, to determine the alpha matte through sampling \cite{sampling_chuang, sampling_feng, sampling_gastal, sampling_he, sampling_johnson, sampling_karacan, sampling_ruzon,yang2018active} or propagation \cite{prop_aksoy2, prop_aksoy, prop_bai, prop_chen, prop_grady, prop_levin, prop_levin2, prop_sun}, which often fail in complex scenes. With the tremendous progress of deep learning, many methods based on CNNs have been proposed with notable successes. Cho {\it et al.} \cite{NIMUDCNN} and Shen {\it et al.} \cite{DAPM} combined the classic algorithms with CNNs for alpha matte refinement. Xu {\it et al.} \cite{DIM} proposed an auto-encoder architecture to predict an alpha matte from a RGB image and a trimap. Some works \cite{GCA, IndexMatter} used attention mechanisms to help improve matting performances. 
Cai {\it et al.} \cite{AdaMatting} suggested a trimap refinement process before matting and showed the advantages of using the refined trimaps. 

Since creating trimaps requires users' additional efforts and the quality of the resulting mattes depends on users' experiences, some recent methods (including our MODNet) attempt to avoid them, as described below.

\begin{figure*}[t]
\begin{center}
   \includegraphics[width=0.99\linewidth]{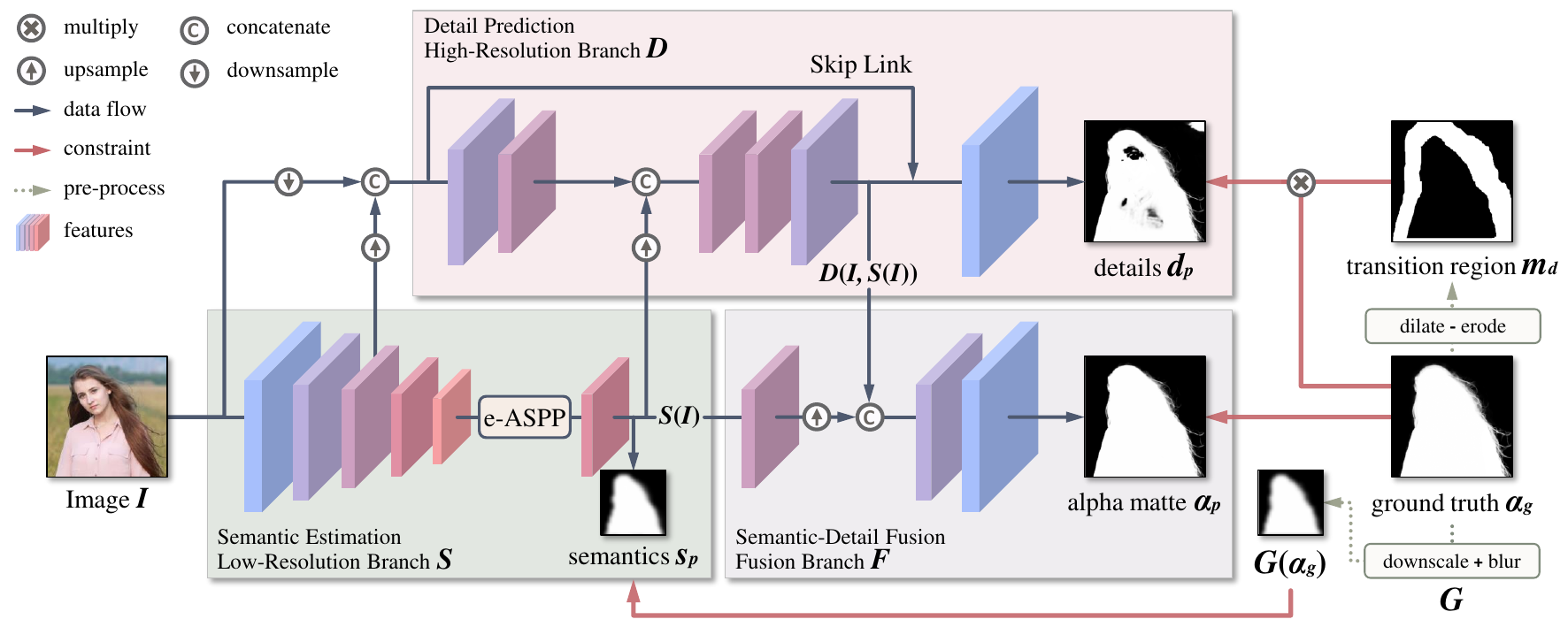}
\end{center}
   \caption{\textbf{Architecture of MODNet.}  
    Given an input image $I$, MODNet predicts portrait semantics $s_p$, boundary details $d_p$, and final alpha matte $\alpha_p$ through three interdependent branches, $S$, $D$, and $F$, which are constrained by explicit supervisions generated from the ground truth matte $\alpha_g$. 
    Since the decomposed sub-objectives are correlated and help strengthen each other, we can optimize MODNet end-to-end.
   }
\label{fig:modnet}
\end{figure*}

\subsection{Trimap-Free Portrait Matting}
Image matting is extremely difficult if trimaps are not available, as a semantic estimation step will then be needed to locate the foreground, before an alpha matte can be predicted. To constrain the problem, most trimap-free methods focus on just one type of foreground objects, {\it e.g.}, portraits. Nevertheless, feeding RGB images to a single network often produce unsatisfactory alpha mattes. Sengupta {\it et al.} \cite{BM} proposed to capture a less expensive background image as a pseudo green screen to alleviate this problem. Other works designed their pipelines with multiple stages. For example, Shen {\it et al.} \cite{SHM} assembled a trimap generation network before the matting network. Zhang {\it et al.} \cite{LFM} applied a fusion network to combine the predicted foreground and background. Liu {\it et al.} \cite{BSHM} concatenated three networks to utilize coarse labeled data in matting. The main problem of all these methods is that they cannot be used in interactive applications as: (1) the background images may change across frames, and (2) computationally expensive due to having multiple stages in the pipeline.
Compared to the above methods, MODNet is light-weight in terms of pipeline complexity. It takes only one RGB image as input and uses a single model for real-time matting with better performances, by optimizing a series of sub-objectives simultaneously with explicit constraints.

\subsection{Other Techniques}
We briefly discuss two techniques that are relevant to the design and optimization of our method.


\noindent\textbf{Atrous Spatial Pyramid Pooling (ASPP).} ASPP~\cite{chen2017deeplab} has been widely explored and proved to boost the performance notably in segmentation-based tasks. Although ASPP has a huge number of parameters and a high computational overhead, some matting models \cite{HAtt, GCA} still used it for better results. In MODNet, we design an efficient variant of ASPP that gives comparable performances with a much lower overhead.

\noindent\textbf{Consistency Constraint.} Consistency is an important assumptions behind many semi-/self-supervised \cite{semi_un_survey} and domain adaptation \cite{udda_survey} algorithms. For example, Ke {\it et al.} \cite{GCT} designed a consistency-based framework that could be used for semi-supervised matting. Toldo {\it et al.} \cite{udamss}  presented a consistency-based domain adaptation strategy for semantic segmentation. However, these methods consist of multiple models and constrain the consistency among their predictions. In contrast, MODNet imposes consistency among various sub-objectives within a model.
\section{MODNet}\label{sec:3}
In MODNet, we divide the trimap-free matting objective into three parts: semantic estimation, detail prediction, and semantic-detail fusion. We optimize them simultaneously via three branches (Fig.~\ref{fig:modnet}). 
In the following subsections, we will delve into the design of each branch and the supervisions used to solve the sub-objectives.

\subsection{Semantic Estimation}\label{sec:3_2}
Similar to existing multi-model approaches, the first step of MODNet is to locate the portrait in the input image $I$. The difference is that we extract high-level semantics only through an encoder, {\it i.e.}, the low-resolution branch $S$ of MODNet. This has two main advantages. First, semantic estimation becomes more efficient because a separate decoder with huge parameters is no longer required. Second, the high-level representation $S(I)$ is helpful for subsequent branches and joint optimization. 
An arbitrary CNN backbone can be applied to $S$.
To facilitate real-time interaction, we adopt MobileNetV2 \cite{net_mobilenetv2}, which is an ingenious model developed for mobile devices, as $S$.

To predict coarse semantic mask $s_p$, we feed $S(I)$ into a convolutional layer activated by the Sigmoid function to reduce its channel number to $1$. We supervise $s_p$ by a thumbnail of the ground truth matte $\alpha_{g}$. Since $s_p$ is supposed to be smooth, we use the L2 loss as:  
\begin{equation}\label{eq:Ls}
    \mathcal{L}_{s} = \frac{1}{2} \, \big|\big| s_p - G(\alpha_g) \big|\big|_{2} \, , 
\end{equation}
where $G$ stands for $16\times$ downsampling followed by Gaussian blur. It removes the fine structures (such as hair) that are not essential to portrait semantics. 

\begin{figure}[t]
\begin{center}
  \includegraphics[width=0.99\linewidth]{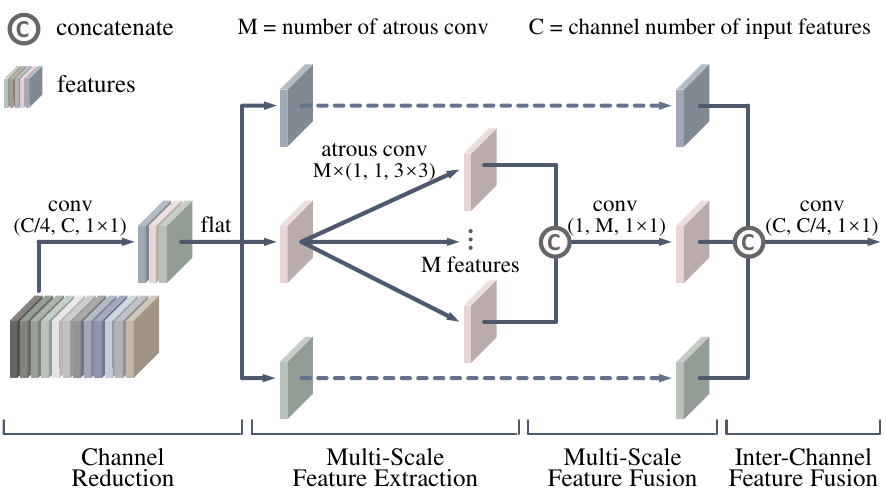}
\end{center}
  \caption{\textbf{Illustration of e-ASPP.} Our e-ASPP is efficient since it extracts and fuses multi-scale features depth-wise, followed by an inter-channel fusion. 
  The tuple under convolution are (output channel, input channel, kernel size). The dotted lines indicate the same structure as the solid line in the center branch.
  }
\label{fig:eASPP}
\end{figure}

\noindent\textbf{Efficient ASPP (e-ASPP).}
Semantic masks predicted by MobileNetV2 may have holes in some challenging foregrounds or backgrounds. Many prior works showed that ASPP was a feasible solution for improving such erroneous semantics. However, ASPP has a very high computational overhead. To balance between performance and efficiency, we design an efficient ASPP (e-ASPP) module to process $S(I)$, as illustrated in Fig.~\ref{fig:eASPP}.

The standard ASPP utilizes atrous convolutions for multi-scale feature extraction and applies a standard convolution for multi-scale feature fusion.
We modify it to e-ASPP via three steps. 
First, we split each atrous convolution into a depth-wise atrous convolution and a point-wise convolution. The depth-wise atrous convolution extracts multi-scale features independently on each channel, while the point-wise convolution is appended for inter-channel fusion at each scale.
Second, we switch the order of inter-channel fusion and the multi-scale feature fusion. In this way,  (1) only one inter-channel fusion is required, and (2) the multi-scale feature fusion is converted to a cheaper depth-wise operation.
Third, we compress the number of input channels by $4\times$ for e-ASPP and recover it before the output. 

Compared to the original ASPP, our proposed e-ASPP has only $1\%$ of the parameters and $1\%$ of the computational overhead\,\footnote{Refer to Appendix A for more details of e-ASPP.}. In MODNet, our experiments show that e-ASPP can achieve performance comparable to ASPP.

\subsection{Detail Prediction}\label{sec:3_3}
We process a transition region around the foreground portrait with a high-resolution branch $D$, which takes $I$, $S(I)$, and the low-level features from $S$ as inputs. The purpose of reusing the low-level features is to reduce the computational overhead of $D$. In addition, we further simplify $D$ in the following three aspects: (1) $D$ consists of fewer convolutional layers than $S$; (2) a small channel number is chosen for the convolutional layers in $D$; (3) we do not maintain the original input resolution throughout $D$. In practice, $D$ consists of $12$ convolutional layers, and its maximum channel number is $64$. The resolution of the feature maps is reduced to $1/4$ of $I$ in the first layer and restored in the last two layers. 
The impact of the downsampling operation on $D$ is negligible since it contains a skip link.

We denote the outputs of $D$ as $D(I, S(I))$, which implies the dependency between sub-objectives --- high-level portrait semantics $S(I)$ is a priori for detail prediction. We calculate the boundary detail matte $d_p$ from $D(I, S(I))$ and learn it through the L1 loss as: 
\begin{equation}
    \mathcal{L}_{d} =  m_d \, \big|\big| d_p - \alpha_g \big|\big|_{1} \, ,
\end{equation}
where $m_d$ is a binary mask to let $\mathcal{L}_d$ focus on the portrait boundaries. $m_d$ is generated through dilation and erosion on $\alpha_{g}$. Its values are $1$ if the pixels are inside the transition region, and $0$ otherwise. 

\subsection{Semantic-Detail Fusion}\label{sec:3_4}
The fusion branch $F$ in MODNet is a straightforward CNN module, combining semantics and details.
We first upsample $S(I)$ to match its size with $D(I, S(I))$. We then concatenate $S(I)$ and $D(I, S(I))$ to predict the final alpha matte $\alpha_{p}$, constrained by:
\begin{equation}
    \mathcal{L}_{\alpha} = \big|\big| \alpha_{p} - \alpha_{g} \big|\big|_{1} + \mathcal{L}_{c} \, , 
\end{equation}
where $\mathcal{L}_c$ is the compositional loss from \cite{DIM}. 
It measures the absolute difference between input image $I$ and the composited image obtained from $\alpha_{p}$, the ground truth foreground, and the ground truth background.

MODNet is trained end-to-end through the sum of $\mathcal{L}_s$, $\mathcal{L}_d$, and $\mathcal{L}_{\alpha}$, as:
\begin{equation}
    \mathcal{L} = \lambda_s \, \mathcal{L}_{s} + \lambda_d \, \mathcal{L}_{d} + \lambda_{\alpha} \, \mathcal{L}_{\alpha} \, , 
\end{equation}
where $\lambda_s$, $\lambda_d$, and $\lambda_{\alpha}$ are hyper-parameters balancing the three losses. The training process is robust to these hyper-parameters. We set $\lambda_s = \lambda_{\alpha} = 1$ and $\lambda_d = 10$.

\begin{figure*}[ht]
\begin{center}
  \includegraphics[width=0.99\linewidth]{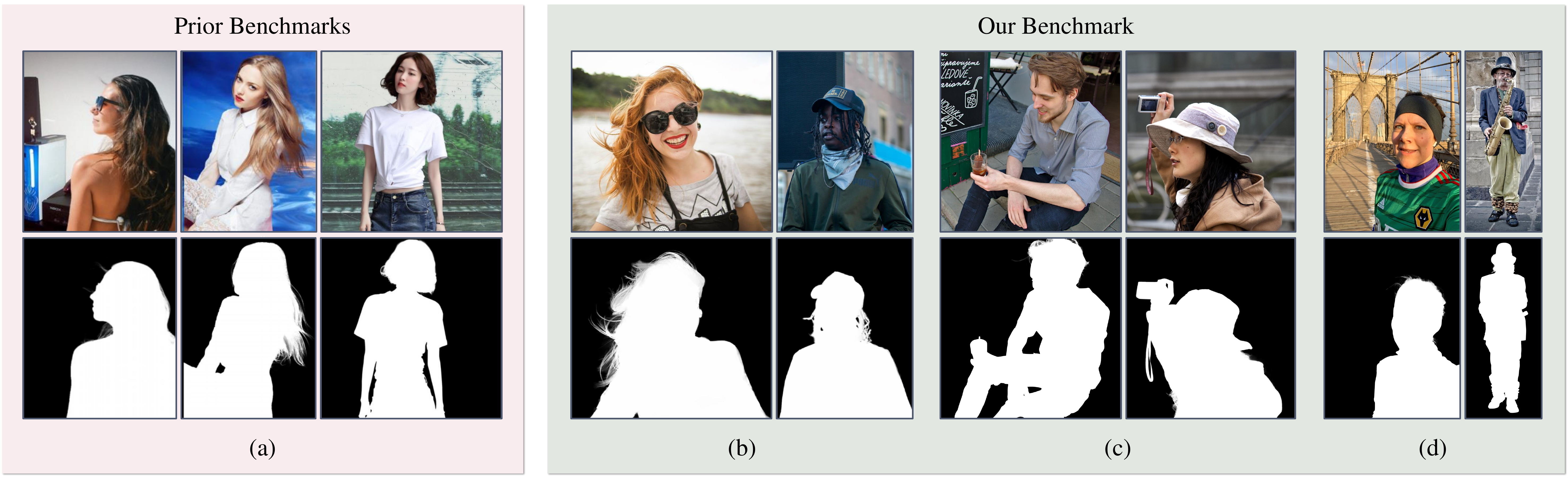}
\end{center}
  \caption{\textbf{Benchmark Comparison.} 
   (a) Validation benchmarks used in \cite{SHM, BSHM, LFM}. Images are synthesized by replacing the original backgrounds with new ones. Instead, our PPM-100 contains original image backgrounds and has a higher diversity in the foregrounds, {\it e.g.}, (b) with fine hairs, (c) with additional objects, and (d) without bokeh or with full-body.
  }
\label{fig:benchmark}
\end{figure*}

\section{SOC for Real-World Data}\label{sec:soc}

The training data for portrait matting requires excellent labeling in the hair area, which is difficult to do for natural images with complex backgrounds. Currently, most annotated data comes from photography websites. Although these images have monochromatic or blurred backgrounds, the labeling process still needs to be completed by experienced annotators with considerable amount of time. As such, the labeled datasets for portrait matting are usually small. Xu {\it et al.} \cite{DIM} suggested using background replacement as a data augmentation to enlarge the training set, and it has become a common setting in image matting. However, the training samples obtained in such a way exhibit different properties from those of the daily life images. 
Therefore, existing trimap-free models always tend to overfit the training set and perform poorly on real-world data. 

To address this domain shift problem, we propose to utilize sub-objectives consistency (SOC) to adapt MODNet to unseen data distributions. The three sub-objectives in MODNet should have consistent outputs in semantics or details. We take semantic consistency as an example, MODNet outputs portrait semantics $S(I)$ and alpha matte $F(S(I), D(S(I)))$ for input image $I$. As $S(I)$ is the prior of $F(S(I), D(S(I)))$, they should have consistent portrait semantics. In the labeled source domain, there is good consistency among the predictions of sub-objectives. However, inconsistent predictions occur in the unlabeled target domain, which may cause poor results. Motivated by this observation, our self-supervised SOC strategy imposes the consistency constraints among the predictions of the sub-objectives (Fig.~\ref{fig:framework}(b)) to improve the performance of MODNet in the new domain, without ground truth labels.

Formally, we denote MODNet as $M$. As described in Sec.~\ref{sec:3}, $M$ has three outputs for an unlabeled image $\tilde{I}$:
\begin{equation}
    \tilde{s}_{p}, \, \tilde{d}_{p}, \, \tilde{\alpha}_{p} = M(\tilde{I}) \, .
\end{equation}
We enforce the semantics in $\tilde{\alpha}_{p}$ to be consistent with $\tilde{s}_{p}$ and the details in $\tilde{\alpha}_{p}$ to be consistent with $\tilde{d}_{p}$ by:
\begin{equation}\label{eq:Lcons}
    \mathcal{L}_{cons} = \frac{1}{2} \, \big|\big| G(\tilde{\alpha}_{p}) - \tilde{s}_{p} \big|\big|_{2}  +  \tilde{m}_d \, \big|\big| \tilde{\alpha}_{p} - \tilde{d}_{p} \big|\big|_{1}  \, ,
\end{equation}
where $\tilde{m}_d$ indicates the transition region in $\tilde{\alpha}_p$, and $G$ has the same meaning as the one in Eq.\,\ref{eq:Ls}.
However, adding the L2 loss to blurred $G(\tilde{\alpha}_{p})$ will smooth the boundaries in the optimized $\tilde{\alpha}_{p}$.
As a result, the consistency between $\tilde{\alpha}_{p}$ and $\tilde{d}_{p}$ will remove the details predicted by the high-resolution branch. To prevent this problem, we duplicate $M$ to $M^{\prime}$ and fix the weights of $M^{\prime}$ before performing SOC. Since the fine boundaries are preserved in $\tilde{d}^{\prime}_p$ output by $M^{\prime}$, we append an extra regularization term to maintain the details in $M$ as:
\begin{equation}\label{eq:Ldc}
    \mathcal{L}_{dd} = \tilde{m}_d \, \big|\big| \tilde{d}_{p}^{\prime} - \tilde{d}_{p} \big|\big|_{1} \, .
\end{equation}
The sum of $\mathcal{L}_{cons}$ and $\mathcal{L}_{dd}$ is optimized during SOC.

\section{Experiments}
In this section, we first introduce our PPM-100 benchmark for portrait matting. We then compare MODNet with existing matting methods on both Adobe Matting Dataset (AMD) \cite{DIM} and our PPM-100. We further conduct ablation experiments to evaluate various components of MODNet. Finally, we demonstrate the effectiveness of SOC in adapting MODNet to real-world data.

\subsection{Photographic Portrait Matting Benchmark}\label{sec:benchmark}
Existing works constructed their validation benchmarks from a small amount of labeled data through image synthesis. Their benchmarks are relatively easy due to unnatural fusion or mismatched semantics between the foreground and the background (Fig.~\ref{fig:benchmark}(a)). Hence, trimap-free models may have comparable performances to the trimap-based models on these benchmarks, but unsatisfactory performances on natural images, {\it i.e.}, images without background replacement. This indicates that the performances of trimap-free methods have not been accurately assessed. 

In contrast, we propose a Photographic Portrait Matting benchmark (PPM-100), which contains 100 finely annotated portrait images with various backgrounds. To guarantee sample diversity, we consider several factors in order to balance the sample types in PPM-100, including: (1) whether the whole portrait body is included; (2) whether the image background is blurred; and (3) whether the person is holding additional objects. We regard small objects held by a foreground person as a part of the foreground, which is more in line with practical applications. 
As shown in Fig.~\ref{fig:benchmark}(b)(c)(d), the samples in PPM-100 have more natural backgrounds and richer postures. Hence, PPM-100 can be considered as a more comprehensive benchmark.

\begin{figure*}[ht]
\begin{center}
  \includegraphics[width=0.99\linewidth]{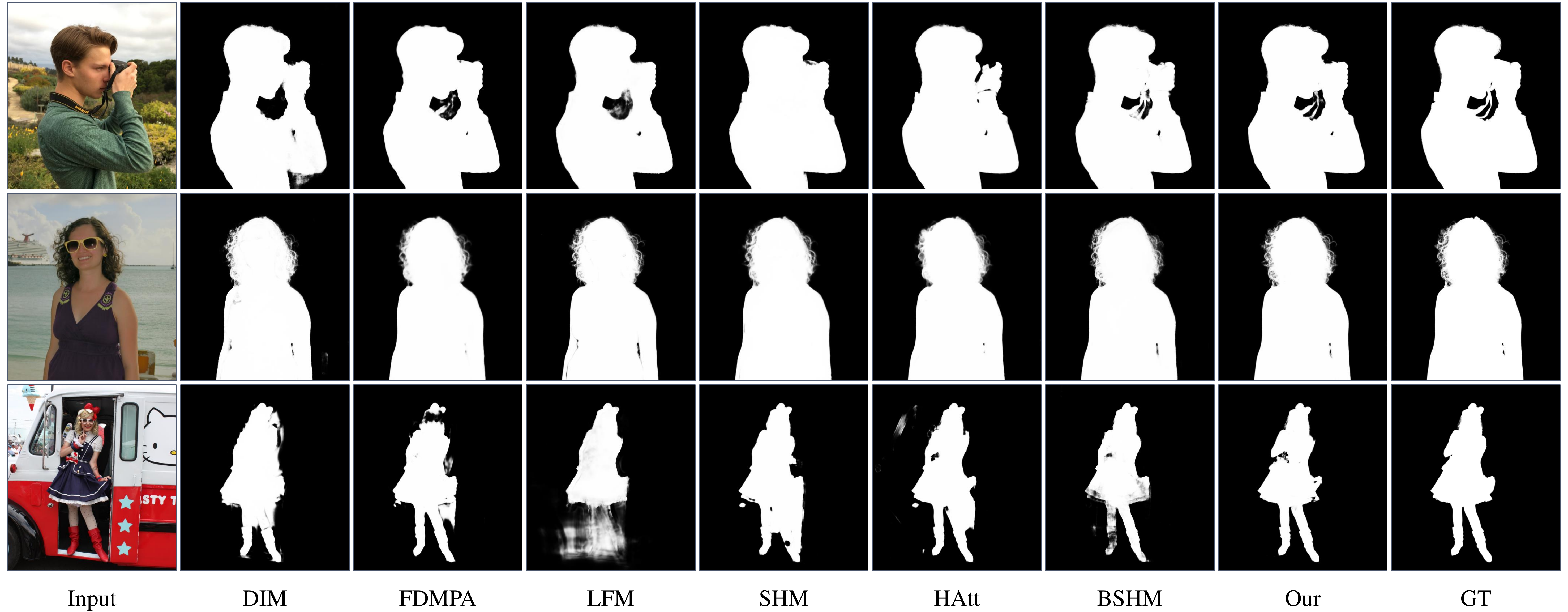}
\end{center}
  \caption{\textbf{Visual Comparison of Trimap-free Methods on PPM-100.}
    MODNet performs better in hollow structures (the 1$st$ row) and hair details (the 2$nd$ row). However, it may still make mistakes in challenging poses or costumes (the 3$rd$ row). DIM \cite{DIM} here does not take trimaps as input, but is pre-trained on the SPS \cite{SPS} dataset. 
  }
\label{fig:visual_results}
\end{figure*}

\subsection{Results on AMD and PPM-100\footnote{Refer to Appendix B for results on more benchmarks.}}\label{sec:benchmark_results}
We compare MODNet with trimap-free FDMPA \cite{FDMPA}, LFM \cite{LFM}, SHM \cite{SHM}, BSHM \cite{BSHM}, and HAtt \cite{HAtt}. We use DIM \cite{DIM} and IndexMatter \cite{IndexMatter} as the trimap-based baselines. For methods without publicly available codes, we follow their papers to reproduce them.

\begin{table}[t]\small
  \begin{center}
     \begin{tabular}{lc|P{2.3}P{2.3}}
      \toprule 
      Method & Trimap & \multicolumn{1}{c}{$\text{MSE} \downarrow$} & \multicolumn{1}{c}{$\text{MAD} \downarrow$} \\
      \midrule
      DIM \cite{DIM} & \checkmark & $0.0016$ & $0.0067$ \\
      IndexMatter \cite{IndexMatter} & \checkmark & $0.0015$ & $0.0064$ \\
      MODNet (Our) & \checkmark & $\textbf{0.0013}$ & $\textbf{0.0054}$ \\
      
      \hline
      DIM \cite{DIM} & & $0.0221$ & $0.0327$ \\
      DIM$^{\dag}$ \cite{DIM} & & $0.0115$ & $0.0178$ \\
      FDMPA$^{\dag}$ \cite{FDMPA} & & $0.0101$ & $0.0160$ \\
      LFM$^{\dag}$ \cite{LFM}  & & $0.0094$ & $0.0158$ \\
      SHM$^{\dag}$ \cite{SHM} & & $0.0072$ & $0.0152$ \\
      HAtt$^{\dag}$ \cite{HAtt} & & $0.0067$ & $0.0137$ \\
      BSHM$^{\dag}$ \cite{BSHM} & & $0.0063$ & $0.0114$ \\
      MODNet$^{\dag}$ (Our) & & $\textbf{0.0044}$ & $\textbf{0.0086}$ \\
      \bottomrule
    \end{tabular}
  \end{center}
  \caption{\textbf{Quantitative Results on PPM-100.} A `$\dag$' indicates that the model is pre-trained on SPS. 
  }
 \label{tab:exp}
\end{table}

\begin{table}[t]\small
  \begin{center}
     \begin{tabular}{lc|P{2.3}P{2.3}}
      \toprule 
      Method & Trimap & \multicolumn{1}{c}{$\text{MSE} \downarrow$} & \multicolumn{1}{c}{$\text{MAD} \downarrow$} \\
      \midrule
      DIM \cite{DIM} & \checkmark & $0.0014$ & $0.0069$ \\
      IndexMatter \cite{IndexMatter} & \checkmark & $0.0013$ & $0.0066$ \\
      MODNet (Our) & \checkmark & $\textbf{0.0011}$ & $\textbf{0.0063}$ \\
      \hline
      DIM \cite{DIM} & & $0.0075$ & $0.0159$ \\
      DIM$^{\dag}$ \cite{DIM} & & $0.0048$ & $0.0116$ \\
      FDMPA$^{\dag}$ \cite{FDMPA} & & $0.0047$ & $0.0115$ \\
      LFM$^{\dag}$ \cite{LFM}  & & $0.0043$ & $0.0101$ \\
      SHM$^{\dag}$ \cite{SHM} & & $0.0031$ & $0.0092$ \\
      HAtt$^{\dag}$ \cite{HAtt} & & $0.0034$ & $0.0094$ \\
      BSHM$^{\dag}$ \cite{BSHM} & & $0.0029$ & $0.0088$ \\
      MODNet$^{\dag}$ (Our) & & $\textbf{0.0023}$ & $\textbf{0.0077}$ \\
      \bottomrule
    \end{tabular}
  \end{center}
  \caption{\textbf{Quantitative Results on AMD.} We pick the portrait foregrounds from AMD for validation. A `$\dag$' indicates that the models pre-trained on SPS.}
 \label{tab:supp_exp}
\end{table}

For a fair comparison, we train all models on the same dataset, which contains nearly $3,000$ annotated foregrounds. Background replacement \cite{DIM} is applied to extend our training set. 
All images in our training set are collected from Flickr and are annotated by Photoshop. The  training set contains $\sim$$2,600$ half-body and $\sim$$400$ full-body portraits. 
For each labeled foreground, we generate $5$ samples by random cropping and $10$ samples by compositing with the images from the OpenImage dataset \cite{openimage} (as the background).
We use MobileNetV2 pre-trained on the Supervisely Person Segmentation (SPS) \cite{SPS} dataset as the backbone of all trimap-free models. For the compared methods, we explore the optimal hyper-parameters through grid search. For MODNet, we train it by SGD for $40$ epochs. With a batch size of $16$, the initial learning rate is set to $0.01$ and is multiplied by $0.1$ after every $10$ epochs. We use Mean Square Error (MSE) and Mean Absolute Difference (MAD) as quantitative metrics.

Table~\ref{tab:exp} shows the results on PPM-100. MODNet outperforms other trimap-free methods on both MSE and MAD. However, it is unable to outperform trimap-based methods, as PPM-100 contains samples with very challenging poses and costumes. 
When taking a trimap as input during both training and testing stages, {\it i.e.}, regarding MODNet as a trimap-based methods,
it outperforms the compared trimap-based methods.
This demonstrates the superiority of the proposed architecture. Fig.~\ref{fig:visual_results} shows visual comparison\,\footnote{Refer to Appendix C for more visual results.}.

\begin{figure}[t]
\begin{center}
  \includegraphics[width=0.99\linewidth]{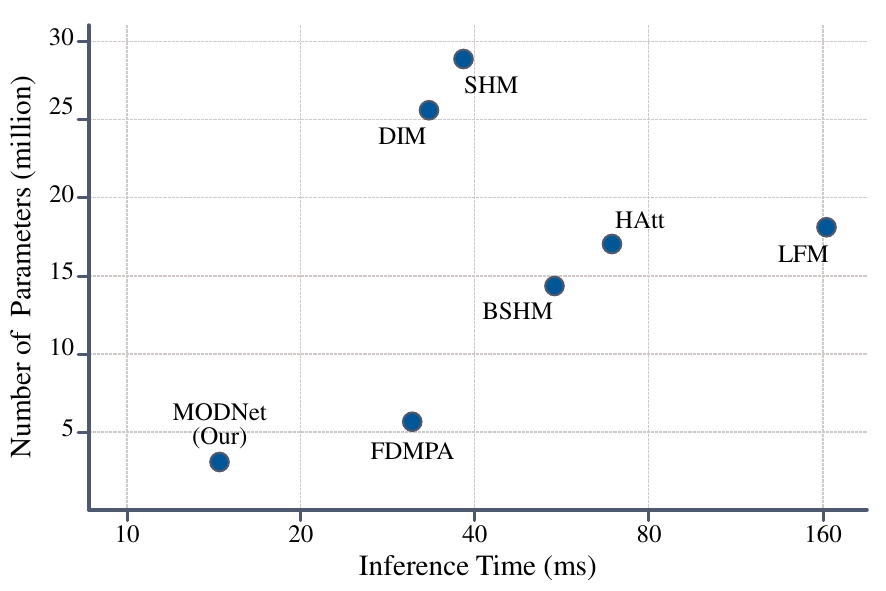}
\end{center}
  \caption{\textbf{Comparison on Model Size and Execution Efficiency.} 
  $fps$ can be obtained by dividing $1,000$ with the inference time.
  }
\label{fig:efficient}
\end{figure}

Table~\ref{tab:supp_exp} shows the results on AMD~\cite{DIM}. We pick the portrait foregrounds from AMD and composite 10 test samples for each foreground with diverse backgrounds. We validate all trained models on this synthetic benchmark. Unlike the results on PPM-100, the performance gap between trimap-free and trimap-based models is much smaller. 
The results show that trimap-free models can achieve results comparable to trimap-based models only on the synthetic benchmarks that have unnatural fusion or mismatched semantics between foreground and background.

We further evaluate MODNet on model size and execution efficiency. A small model facilitates deployment on mobile/handheld devices, while high execution efficiency is necessary for real-time applications. We measure the model size by the total number of parameters, and we reflect the execution efficiency by the average inference time over PPM-100 on an NVIDIA GTX 1080Ti GPU (all input images are resized to $512\times512$). Note that fewer parameters do not imply faster inference speed due to large feature maps or time-consuming mechanisms, {\it e.g.}, attention, that the model may use. Fig.~\ref{fig:efficient} summarizes the results. The inference time of MODNet is $14.9\,ms$ ($67\,fps$), which is twice the $fps$ of the fastest method, FDMPA ($31\,fps$). In addition, our MODNet has the smallest number of parameters among the trimap-free methods.

\begin{table}[t]\small
  \begin{center}
     \begin{tabular}{cccc|P{2.3}P{2.3}}
      \toprule 
      $\mathcal{L}_s$ & $\mathcal{L}_d$ & e-ASPP & SPS & \multicolumn{1}{c}{$\text{MSE}\downarrow$} & \multicolumn{1}{c}{$\text{MAD} \downarrow$} \\
      \midrule
      & & & & $0.0162$ & $0.0235$ \\
      \checkmark & & & & $0.0097$ & $0.0158$ \\
      \checkmark & \checkmark & & & $0.0083$ & $0.0142$ \\
      \checkmark & \checkmark & \checkmark & & $0.0057$ & $0.0109$ \\
      \checkmark & \checkmark & \checkmark & \checkmark & $\textbf{0.0044}$ & $\textbf{0.0086}$ \\
      \bottomrule
    \end{tabular}
  \end{center}
  \caption{\textbf{Ablation of MODNet on PPM-100.} SPS indicates the model us pre-trained on SPS.}
      \label{tab:ablation}
\end{table}

We have also conducted ablation experiments for MODNet on PPM-100, as shown in  Table~\ref{tab:ablation}. Applying $\mathcal{L}_{s}$ and $\mathcal{L}_{d}$ to constrain portrait semantics and boundary details  bring  considerable performance improvements. The results also show that the effectiveness of e-ASPP in fusing multi-level feature maps. Although SPS pre-training is optional to MODNet, it plays a vital role in other trimap-free methods. From Table~\ref{tab:exp}, we can see that trimap-free DIM without pre-training performs far worse than the one with pre-training.

\begin{figure}[t]
\begin{center}
  \includegraphics[width=0.99\linewidth]{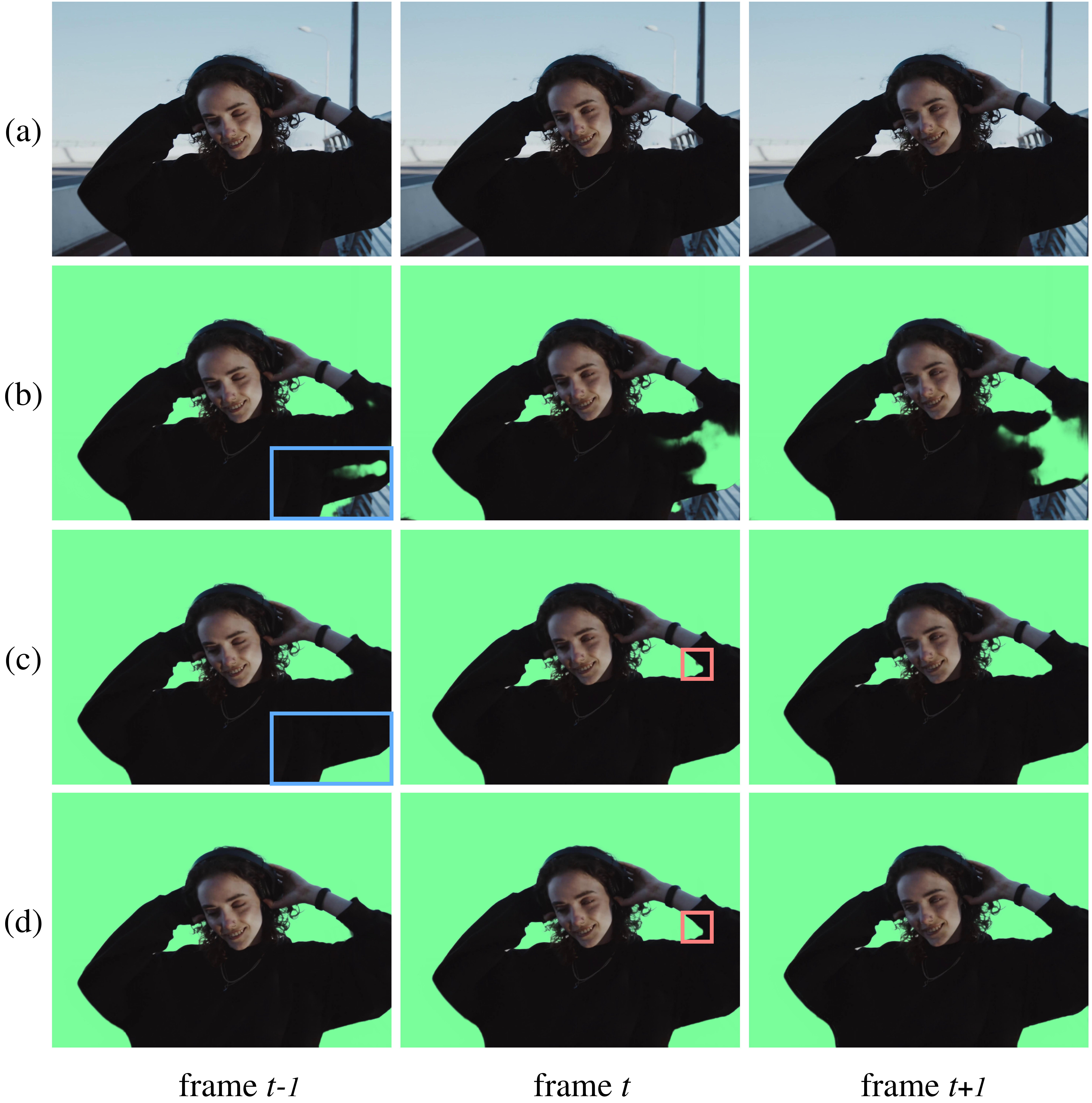}
\end{center}
  \caption{\textbf{Results on a Real-World Video.} 
We show three consecutive video frames from left to right. From top to bottom: (a) Input, (b) MODNet, (c) MODNet + SOC, and (d) MODNet + SOC + OFD. The blue region in frame $t-1$ shows the effectiveness of SOC, while the red region in frame $t$ highlights the flickers eliminated by OFD.
  }
\label{fig:video}
\end{figure}

\subsection{Results on Real-World Data}\label{sec:RoRWD}
To adapt MODNet to real-world data, we capture $\sim$400 video clips (divided into about 50,000 frames) as the unlabeled data for self-supervised SOC domain adaptation. In this stage, we freeze the BatchNorm layers within MODNet and finetune the convolutional layers by Adam at a learning rate of $0.0001$. The total number of fine-tuning epochs are 15. Here, we only provide visual results, 
as ground truth mattes are not available. In Fig.~\ref{fig:video}(b)(c), we composite the foreground over a green screen to emphasize that SOC is vital for generalizing MODNet to real-world data. 


For video data, we also propose here a simple but effective One-Frame Delay (OFD) trick to reduce the flickers in the predicted alpha matte sequence. The idea behind OFD is that we can utilize the preceding and the following frames to fix the flickering pixels, because the corresponding pixels in adjacent frames are likely to be correct.
Suppose that we have three consecutive frames, and their corresponding alpha mattes are $\alpha_{t-1}$, $\alpha_{t}$, and $\alpha_{t+1}$, where $t$ is the frame index. We regard $\alpha_{t}^{i}$ as a flickering pixel if the values of $\alpha_{t-1}^{i}$ and $\alpha_{t+1}^{i}$ are close, and $\alpha_{t}^{i}$ is very different from the values of both $\alpha_{t-1}^{i}$ and $\alpha_{t+1}^{i}$. When $\alpha_{t}^{i}$ is a flickering pixel, we replace its value by averaging $\alpha_{t-1}^{i}$ and $\alpha_{t+1}^{i}$. As illustrated in Fig.~\ref{fig:video}(c)(d), OFD can further removes flickers along the boundaries. 




\section{Conclusion}
This paper has presented a simple, fast, and effective model, MODNet, for portrait matting. By taking only an RGB image as input, our method enables the prediction of a high-quality alpha matte in real time, which is benefited from objective decomposition and concurrent optimization with explicit supervisions. Besides, we have introduced (1) an e-ASPP module to speed up the multi-scale feature fusion process, and (2) a self-supervised sub-objectives consistency (SOC) strategy to allow MODNet to handle the domain shift problem. Extensive experiments show that MODNet outperforms existing trimap-free methods on the AMD benchmark, the proposed PPM-100 benchmark, and a variety of real-world data. Our method does have limitations. The main one is that it may fail to handle videos with strong motion blurs due to the lack of temporal information. One possible future work is to address the video matting problem under motion blurs through additional sub-objectives, such as optical flow estimation. 

\section{Acknowledge}
We thank Yurou Zhou, Qiuhua Wu, and Xiangyu Mao from SenseTime Research for their discussions and help in this work.



\section*{Appendix A: Analysis of e-ASPP}
Here we compare the proposed Efficient ASPP (e-ASPP) with the standard ASPP in terms of the number of parameters and computational overhead. For a convolutional layer, the number of its parameters $\mathcal{P}$ can be calculated by:
\begin{equation}
    \mathcal{P} = C_{out} \times C_{in} \times K \times K,
\end{equation}
where $C_{out}$ is the number of output channels, $C_{in}$ is the number of input channels, and $K$ is the kernel size. We can use {\it FLOPs} to measure the computational overhead $\mathcal{O}$ of a convolutional layer as:
\begin{equation}
    \mathcal{O} = C_{in} \times 2 \times K \times K \times H_{out} \times W_{out} \times C_{out},
\end{equation}
where $H_{out}$ and $W_{out}$ are the height and the width of output feature maps, respectively.

Following, we represent the size of the input feature maps by $(c, h, w)$, where $c$ is the number of channels, $h$ is the height of the input feature maps, and $w$ is the width of the input feature maps. We represent the number of atrous convolutional layers (with a kernel size of $k$) in both ASPP and e-ASPP by $m$.

\textbf{Standard ASPP (ASPP).} In ASPP, (1) all atrous convolutional layers are independently applied to the input features maps to extract multi-scale features. These multi-scale features are then (2) concatenated and processed by a point-wise convolutional layer (with a kernel size of $1$). We have:
\begin{equation}
\begin{split}
        \mathcal{P_{ASPP}} =& m \times (c \times c \times k \times k) \\ &+ c \times (m \times c) \times 1 \times 1 \\
    =& m \times c^2 \times (k^2 + 1), \\
\end{split}
\end{equation}

\begin{equation}
\begin{split}
    \mathcal{O}_{ASPP} =& m \times (c \times 2 \times k \times k \times h \times w \times c) \\ 
    &+ (m \times c) \times 2 \times 1 \times 1 \times h \times w \times c \\
    =& ((2 \times k^2 + 2) \times m \times c) \times (h \times w \times c).
\end{split}
\end{equation}

\textbf{Efficient ASPP (e-ASPP).} As shown in Fig.\,{\color{red} 3} (in the paper), e-ASPP consists of four operations, including (1) Channel Reduction, (2) Multi-Scale Feature Extraction, (3) Multi-Scale Feature Fusion, and (4) Inter-Channel Feature Fusion. The total number of parameters and the total {\it FLOPs} are the sum of these four operations. We have:

\begin{equation}
\begin{split}
        \mathcal{P}_{e-ASPP} =& \frac{c}{4} \times c \times 1 \times 1 \\ &+ \frac{c}{4} \times m \times (1 \times 1 \times k \times k) \\ 
        &+ \frac{c}{4} \times (1 \times m \times 1 \times 1)
        \\ &+ c \times \frac{c}{4} \times 1 \times 1 \\
        =& \frac{2 \times c^2 + (k^2 + 1) \times m \times c}{4}, \\
\end{split}
\end{equation}

\begin{equation}
\begin{split}
        \mathcal{O}_{e-ASPP} =& c \times 2 \times 1 \times 1 \times h \times w \times \frac{c}{4} \\
        &+ \frac{c}{4} \times m \times (1 \times 2 \times k \times k \times h \times w \times 1) \\
        &+ \frac{c}{4} \times (m \times 2 \times 1 \times 1 \times h \times w \times 1) \\
        &+ \frac{c}{4} \times 2 \times 1 \times 1 \times h \times w \times c \\
        =& (c + \frac{(k^2 + 1) \times m}{2}) \times (h \times w \times c) .
\end{split}
\end{equation}

Following the standard ASPP, we set $k = 3$ and $m = 5$. Usually, $c \geq 256$ is applied in most networks. Therefore, we have:
\begin{equation}
\begin{split}
        \frac{\mathcal{P}_{e-ASPP}}{\mathcal{P}_{ASPP}} \approx 0.01,
\end{split}
\end{equation}
\begin{equation}
\begin{split}
        \frac{\mathcal{O}_{e-ASPP}}{\mathcal{O}_{ASPP}} \approx 0.01.
\end{split}
\end{equation}
It means that compared to the standard ASPP, our proposed e-ASPP
has only $1\%$ of the parameters and $1\%$ of the computational
overhead. In MODNet, our experiments show that e-ASPP
can achieve performance comparable to ASPP.
Note that when the Channel Reduction operation in e-ASPP is disabled, e-ASPP still has only $2\%$ of the parameters and $2\%$ of the computational overhead compared to ASPP.

\section*{Appendix B: Results on CRGNN-R and D646}
In Table~\ref{tab:CRGNN-R}, we provide the quantitative results on a video matting dataset proposed by~\cite{CRGNN} to show the effectiveness of the proposed SOC strategy. In Table~\ref{tab:D646}, we compare MODNet with previous SOTA methods on the D646 dataset proposed by~\cite{HAtt}. 

\twocolumn[{%
\renewcommand\twocolumn[1][]{#1}%
\maketitle
\begin{center}\label{fig:mvc}
    \centering
    \includegraphics[width=0.99\linewidth]{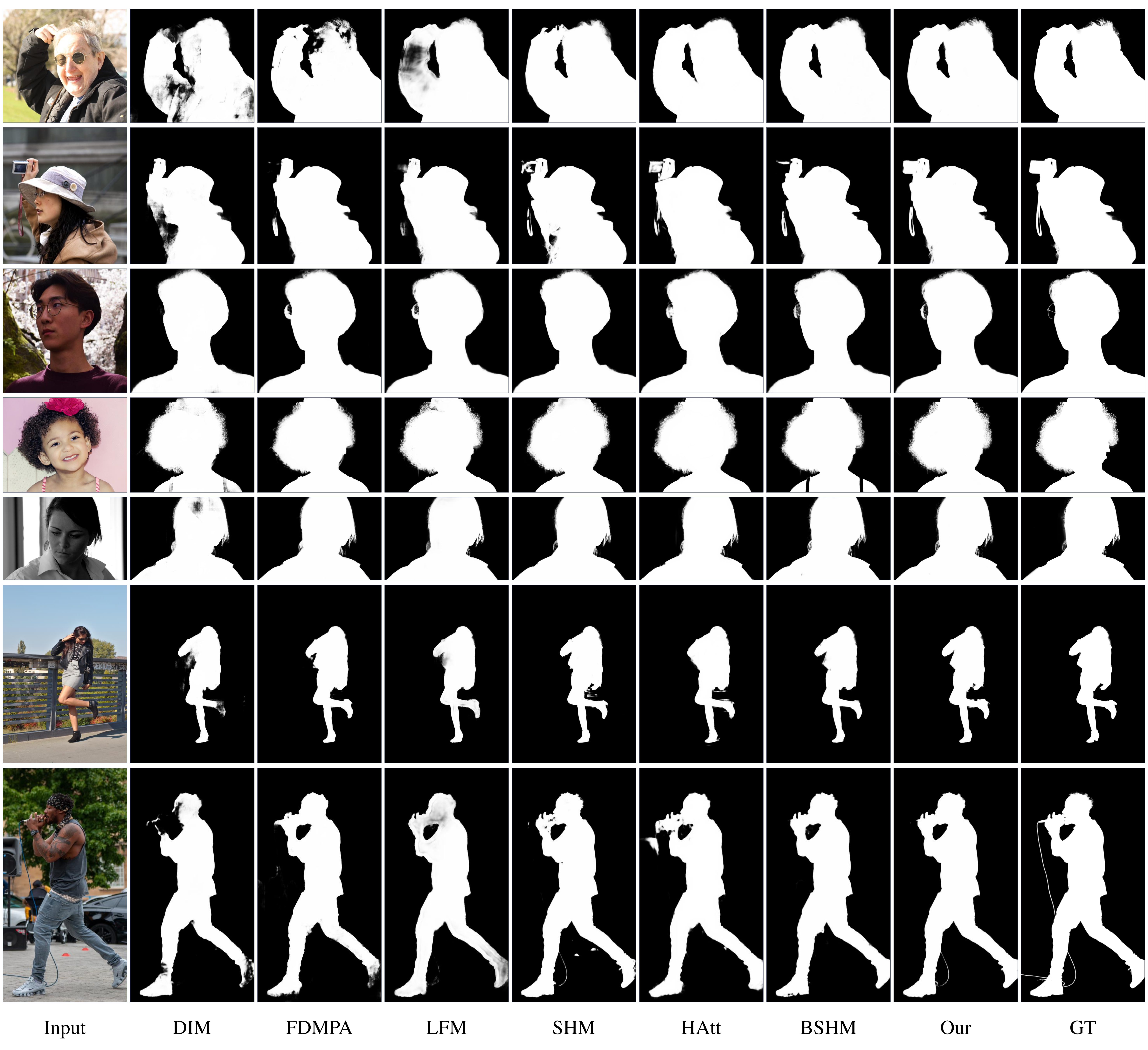}
    \captionof{figure}{\textbf{More Visual Comparisons of Trimap-free Methods on PHM-100.} We compare our MODNet with DIM \cite{DIM}, FDMPA \cite{FDMPA}, LFM \cite{LFM}, SHM \cite{SHM}, HAtt \cite{HAtt}, and BSHM \cite{BSHM}. Note that DIM here does not take trimaps as the input but is pre-trained on the SPS \cite{SPS} dataset. Zoom in for the best visualization.}
\end{center}%
}]


\section*{Appendix C: Visual Results on PHM-100}
Fig.\,8 provides more visual comparisons of MODNet and the existing trimap-free methods on PHM-100.

\begin{table}[t]\small
  \begin{center}
     \begin{tabular}{lc|P{2.3}P{2.3}}
      \toprule 
      Method & Trimap & \multicolumn{1}{c}{$\text{MSE} \downarrow$} & \multicolumn{1}{c}{$\text{MAD} \downarrow$} \\
      \midrule
      CRGNN~\cite{CRGNN} & \checkmark &$0.0010$   &$0.0035$  \\
      \hline
      MODNet (Our) & & $0.0082$ & $0.0157$ \\
      MODNet + SOC (Our) & & $\textbf{0.0033}$ & $\textbf{0.0084}$ \\
      \bottomrule
    \end{tabular}
  \end{center}
   \vspace{-0.3cm}
  \caption{Results on CRGNN-R~\cite{CRGNN}.}
   \vspace{-0.1cm}
  \label{tab:CRGNN-R}
\end{table}

\begin{table}[t]\small
  \begin{center}
     \begin{tabular}{lc|P{2.3}P{2.3}}
      \toprule 
      Method & Trimap & \multicolumn{1}{c}{$\text{MSE} \downarrow$} & \multicolumn{1}{c}{$\text{MAD} \downarrow$} \\
      \midrule
      DIM~\cite{DIM} & \checkmark & $0.0025$ & $0.0081$ \\
      \hline
      HAtt~\cite{HAtt} & & $0.0054$ & $0.0126$ \\
      MODNet (Our) & & $\textbf{0.0037}$ & $\textbf{0.0098}$ \\
      \bottomrule
    \end{tabular}
  \end{center}
  \vspace{-0.3cm}
  \caption{Results on D646~\cite{HAtt}.}
  \vspace{-0.4cm}
    \label{tab:D646}
\end{table}

\section*{Appendix D: Comparison with BM}
\begin{figure}[t]
\begin{center}
  \includegraphics[width=0.99\linewidth]{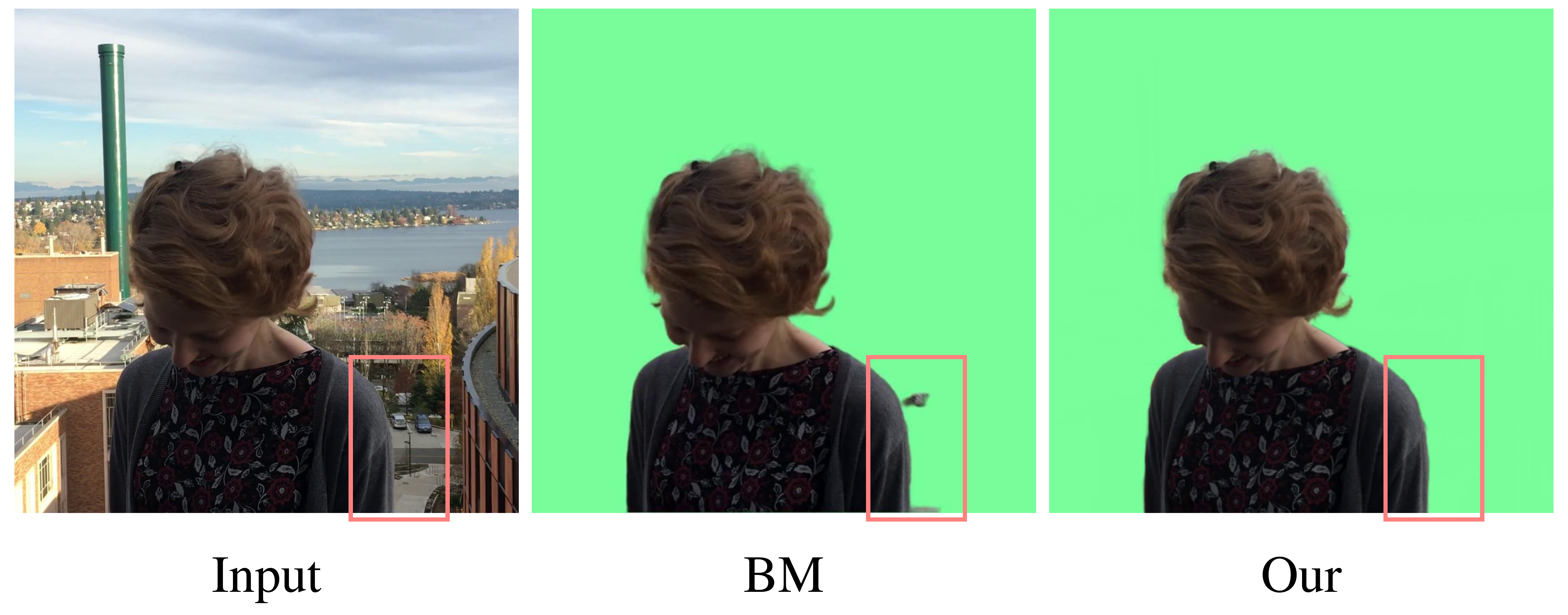}
\end{center}
  \caption{\textbf{MODNet versus BM with a fixed camera position.} MODNet outperforms BM \cite{BM} when a car is entering the background (red region).
  }
\label{fig:bmc}
\end{figure}

We compare MODNet against the background matting (BM) proposed by \cite{BM}. Since BM does not support dynamic backgrounds, we conduct validations in the fixed-camera scenes from \cite{BM}. BM relies on a static background image, which implicitly assumes that all pixels whose value changes across frames belong to the foreground. As shown in Fig.~\ref{fig:bmc}, when a moving object suddenly appears in the background, the result of BM will be affected, but MODNet is robust to such disturbances.

\bibliography{aaai22}

\begin{thebibliography}{41}
\providecommand{\natexlab}[1]{#1}

\bibitem[{Aksoy, Aydin, and Pollefeys(2017)}]{prop_aksoy2}
Aksoy, Y.; Aydin, T.~O.; and Pollefeys, M. 2017.
\newblock Designing effective inter-pixel information flow for natural image
  matting.
\newblock In \emph{CVPR}.

\bibitem[{Aksoy et~al.(2018)Aksoy, Oh, Paris, Pollefeys, and
  Matusik}]{prop_aksoy}
Aksoy, Y.; Oh, T.-H.; Paris, S.; Pollefeys, M.; and Matusik, W. 2018.
\newblock Semantic soft segmentation.
\newblock \emph{TOG}.

\bibitem[{Bai and Sapiro(2007)}]{prop_bai}
Bai, X.; and Sapiro, G. 2007.
\newblock A geodesic framework for fast interactive image and video
  segmentation and matting.
\newblock In \emph{ICCV}.

\bibitem[{Cai et~al.(2019)Cai, Zhang, Fan, Huang, Liu, Liu, Liu, Wang, and
  Sun}]{AdaMatting}
Cai, S.; Zhang, X.; Fan, H.; Huang, H.; Liu, J.; Liu, J.; Liu, J.; Wang, J.;
  and Sun, J. 2019.
\newblock Disentangled Image Matting.
\newblock In \emph{ICCV}.

\bibitem[{Chen et~al.(2017)Chen, Papandreou, Kokkinos, Murphy, and
  Yuille}]{chen2017deeplab}
Chen, L.-C.; Papandreou, G.; Kokkinos, I.; Murphy, K.; and Yuille, A.~L. 2017.
\newblock Deeplab: Semantic image segmentation with deep convolutional nets,
  atrous convolution, and fully connected crfs.
\newblock \emph{PAMI}, 40(4): 834--848.

\bibitem[{Chen et~al.(2018)Chen, Ge, Xu, Zhang, Yang, and Gai}]{SHM}
Chen, Q.; Ge, T.; Xu, Y.; Zhang, Z.; Yang, X.; and Gai, K. 2018.
\newblock Semantic human matting.
\newblock In \emph{ACMMM}.

\bibitem[{Chen, Li, and Tang(2013)}]{prop_chen}
Chen, Q.; Li, D.; and Tang, C.-K. 2013.
\newblock KNN Matting.
\newblock \emph{PAMI}.

\bibitem[{Cho, Tai, and Kweon(2016)}]{NIMUDCNN}
Cho, D.; Tai, Y.-W.; and Kweon, I. 2016.
\newblock Natural image matting using deep convolutional neural networks.
\newblock In \emph{ECCV}.

\bibitem[{Chuang et~al.(2001)Chuang, Curless, Salesin, and
  Szeliski}]{sampling_chuang}
Chuang, Y.-Y.; Curless, B.; Salesin, D.~H.; and Szeliski, R. 2001.
\newblock A bayesian approach to digital matting.
\newblock In \emph{CVPR}.

\bibitem[{Feng, Liang, and Zhang(2016)}]{sampling_feng}
Feng, X.; Liang, X.; and Zhang, Z. 2016.
\newblock A cluster sampling method for image matting via sparse coding.
\newblock In \emph{ECCV}.

\bibitem[{Foix, Alenyà, and Torras(2011)}]{ToF}
Foix, S.; Alenyà, G.; and Torras, C. 2011.
\newblock Lock-in Time-of-Flight (ToF) cameras: A survey.
\newblock \emph{Sensors Journal}.

\bibitem[{Gastal and Oliveira(2010)}]{sampling_gastal}
Gastal, E. S.~L.; and Oliveira, M.~M. 2010.
\newblock Shared sampling for real-time alpha matting.
\newblock In \emph{Eurographics}.

\bibitem[{Grady et~al.(2005)Grady, Schiwietz, Aharon, and
  Westermann}]{prop_grady}
Grady, L.; Schiwietz, T.; Aharon, S.; and Westermann, R. 2005.
\newblock Random walks for interactive alpha-matting.
\newblock In \emph{VIIP}.

\bibitem[{He et~al.(2011)He, Rhaemann, Rother, Tang, and Sun}]{sampling_he}
He, K.; Rhaemann, C.; Rother, C.; Tang, X.; and Sun, J. 2011.
\newblock A global sampling method for alpha matting.
\newblock In \emph{CVPR}.

\bibitem[{Hou and Liu(2019)}]{CAMatting}
Hou, Q.; and Liu, F. 2019.
\newblock Context-aware Image Matting for Simultaneous Foreground and Alpha
  Estimation.
\newblock In \emph{ICCV}.

\bibitem[{Johnson et~al.(2016)Johnson, Varnousfaderani, Cholakkal, and
  Rajan}]{sampling_johnson}
Johnson, J.; Varnousfaderani, E.~S.; Cholakkal, H.; and Rajan, D. 2016.
\newblock Sparse coding for alpha matting.
\newblock \emph{TIP}.

\bibitem[{Karacan, Erdem, and Erdem(2015)}]{sampling_karacan}
Karacan, L.; Erdem, A.; and Erdem, E. 2015.
\newblock Image matting with kl-divergence based sparse sampling.
\newblock In \emph{ICCV}.

\bibitem[{Ke et~al.(2020)Ke, Qiu, Li, Yan, and Lau}]{GCT}
Ke, Z.; Qiu, D.; Li, K.; Yan, Q.; and Lau, R.~W. 2020.
\newblock Guided Collaborative Training for Pixel-wise Semi-Supervised
  Learning.
\newblock In \emph{ECCV}.

\bibitem[{Kuznetsova et~al.(2018)Kuznetsova, Rom, Alldrin, Uijlings, Krasin,
  Pont{-}Tuset, Kamali, Popov, Malloci, Duerig, and Ferrari}]{openimage}
Kuznetsova, A.; Rom, H.; Alldrin, N.; Uijlings, J. R.~R.; Krasin, I.;
  Pont{-}Tuset, J.; Kamali, S.; Popov, S.; Malloci, M.; Duerig, T.; and
  Ferrari, V. 2018.
\newblock The Open Images Dataset V4: Unified image classification, object
  detection, and visual relationship detection at scale.
\newblock \emph{IJCV}.

\bibitem[{Levin, Lischinski, and Weiss(2007)}]{prop_levin}
Levin, A.; Lischinski, D.; and Weiss, Y. 2007.
\newblock A closed-form solution to natural image matting.
\newblock \emph{PAMI}.

\bibitem[{Levin, Rav-Acha, and Lischinski(2008)}]{prop_levin2}
Levin, A.; Rav-Acha, A.; and Lischinski, D. 2008.
\newblock Spectral matting.
\newblock \emph{PAMI}.

\bibitem[{Li and Lu(2020)}]{GCA}
Li, Y.; and Lu, H. 2020.
\newblock Natural image matting via guided contextual attention.
\newblock In \emph{AAAI}.

\bibitem[{Liu et~al.(2020)Liu, Yao, Hou, Cui, Xie, Zhang, and Hua}]{BSHM}
Liu, J.; Yao, Y.; Hou, W.; Cui, M.; Xie, X.; Zhang, C.; and Hua, X.-S. 2020.
\newblock Boosting Semantic Human Matting With Coarse Annotations.
\newblock In \emph{CVPR}.

\bibitem[{Lu et~al.(2019)Lu, Dai, Shen, and Xu}]{IndexMatter}
Lu, H.; Dai, Y.; Shen, C.; and Xu, S. 2019.
\newblock Indices Matter: Learning to Index for Deep Image Matting.
\newblock In \emph{ICCV}.

\bibitem[{Qiao et~al.(2020)Qiao, Liu, Yang, Zhou, Xu, Zhang, and Wei1}]{HAtt}
Qiao, Y.; Liu, Y.; Yang, X.; Zhou, D.; Xu, M.; Zhang, Q.; and Wei1, X. 2020.
\newblock Attention-Guided Hierarchical Structure Aggregation for Image
  Matting.
\newblock In \emph{CVPR}.

\bibitem[{Ruzon and Tomasi(2000)}]{sampling_ruzon}
Ruzon, M.~A.; and Tomasi, C. 2000.
\newblock Alpha estimation in natural images.
\newblock In \emph{CVPR}.

\bibitem[{Sandler et~al.(2018)Sandler, Howard, Zhu, Zhmoginov, and
  Chen}]{net_mobilenetv2}
Sandler, M.; Howard, A.; Zhu, M.; Zhmoginov, A.; and Chen, L.-C. 2018.
\newblock MobileNetV2: Inverted Residuals and Linear Bottlenecks.
\newblock In \emph{CVPR}.

\bibitem[{Schmarje et~al.(2020)Schmarje, Santarossa, Schröder, and
  Koch}]{semi_un_survey}
Schmarje, L.; Santarossa, M.; Schröder, S.-M.; and Koch, R. 2020.
\newblock A survey on Semi-, Self- and Unsupervised Learning for Image
  Classification.
\newblock \emph{ArXiv}, abs/2002.08721.

\bibitem[{Sengupta et~al.(2020)Sengupta, Jayaram, Curless, Seitz, and
  Kemelmacher-Shlizerman}]{BM}
Sengupta, S.; Jayaram, V.; Curless, B.; Seitz, S.; and Kemelmacher-Shlizerman,
  I. 2020.
\newblock Background Matting: The World is Your Green Screen.
\newblock In \emph{CVPR}.

\bibitem[{Shen et~al.(2016)Shen, Tao, Gao, Zhou, and Jia}]{DAPM}
Shen, X.; Tao, X.; Gao, H.; Zhou, C.; and Jia, J. 2016.
\newblock Deep automatic portrait matting.
\newblock In \emph{ECCV}.

\bibitem[{Sun, Feng, and Saenko(2016)}]{DomainShift}
Sun, B.; Feng, J.; and Saenko, K. 2016.
\newblock Return of Frustratingly Easy Domain Adaptation.
\newblock In \emph{AAAI}.

\bibitem[{Sun et~al.(2004)Sun, Jia, Tang, and Shum}]{prop_sun}
Sun, J.; Jia, J.; Tang, C.-K.; and Shum, H.-Y. 2004.
\newblock Poisson matting.
\newblock \emph{TOG}.

\bibitem[{supervise.ly(2018)}]{SPS}
supervise.ly. 2018.
\newblock Supervisely Person Dataset.
\newblock \emph{supervise.ly}.

\bibitem[{Tang et~al.(2019)Tang, Aksoy, Oztireli, Gross, and
  Aydin}]{SampleMatting}
Tang, J.; Aksoy, Y.; Oztireli, C.; Gross, M.; and Aydin, T.~O. 2019.
\newblock Learning-based Sampling for Natural Image Matting.
\newblock In \emph{CVPR}.

\bibitem[{Toldo et~al.(2020)Toldo, Michieli, Agresti, and Zanuttigh}]{udamss}
Toldo, M.; Michieli, U.; Agresti, G.; and Zanuttigh, P. 2020.
\newblock Unsupervised Domain Adaptation for Mobile Semantic Segmentation based
  on Cycle Consistency and Feature Alignment.
\newblock \emph{IMAVIS}.

\bibitem[{Wang et~al.(2021)Wang, Liu, Tian, Li, and Yang}]{CRGNN}
Wang, T.; Liu, S.; Tian, Y.; Li, K.; and Yang, M.-H. 2021.
\newblock Video Matting via Consistency-Regularized Graph Neural Networks.
\newblock In \emph{ICCV}.

\bibitem[{Wilson and Cook(2020)}]{udda_survey}
Wilson, G.; and Cook, D.~J. 2020.
\newblock A Survey of Unsupervised Deep Domain Adaptation.
\newblock \emph{TIST}.

\bibitem[{Xu et~al.(2017)Xu, Price, Cohen, and Huang}]{DIM}
Xu, N.; Price, B.; Cohen, S.; and Huang, T. 2017.
\newblock Deep Image Matting.
\newblock In \emph{CVPR}.

\bibitem[{Yang et~al.(2018)Yang, Xu, Chen, He, Yin, and Lau}]{yang2018active}
Yang, X.; Xu, K.; Chen, S.; He, S.; Yin, B.~Y.; and Lau, R. 2018.
\newblock Active matting.
\newblock \emph{Adv. Neural Inform. Process. Syst.}

\bibitem[{Zhang et~al.(2019)Zhang, Gong, Fan, Ren, Huang, Bao, and Xu}]{LFM}
Zhang, Y.; Gong, L.; Fan, L.; Ren, P.; Huang, Q.; Bao, H.; and Xu, W. 2019.
\newblock A late fusion cnn for digital matting.
\newblock In \emph{CVPR}.

\bibitem[{Zhu et~al.(2017)Zhu, Chen, Wang, Liu, Zhang, and Tang}]{FDMPA}
Zhu, B.; Chen, Y.; Wang, J.; Liu, S.; Zhang, B.; and Tang, M. 2017.
\newblock Fast Deep Matting for Portrait Animation on Mobile Phone.
\newblock In \emph{ACMMM}.

\end{thebibliography}

\end{document}